\documentclass[11pt,a4paper]{article}
\usepackage[hyperref]{naaclhlt2019}
\usepackage{times}
\usepackage{latexsym}
\usepackage{graphicx}  
\usepackage{url}
\usepackage{amsmath}
\usepackage{bm}      
\usepackage{booktabs,subcaption,amsfonts,dcolumn}
\newcolumntype{d}[1]{D..{#1}}
\usepackage{hhline}
\newtheorem{theorem}{Theorem}
\aclfinalcopy 

\usepackage{amssymb}
\usepackage{pifont}
\title{A General FOFE-net Framework for Simple and Effective Question Answering over Knowledge Bases}

\author{
  Dekun Wu\\
  York University\\
  \texttt{jackwu@eecs.yorku.ca}
  \\\And
  Nana Nosirova\\
  University of Toronto\\
  \texttt{nanalelfe@gmail.com}
  \\ \AND
  Hui Jiang\\
  York University\\
  \texttt{hj@eecs.yorku.ca}
  \\\And
  Mingbin Xu\\
  York University\\
  \texttt{mingbin.xu@gmail.com}
}

\date{}

\begin{document}
\maketitle
\begin{abstract}
Question answering over knowledge base (KB-QA) has recently become a popular research topic in NLP. One popular way to solve the KB-QA problem is to make use of a pipeline of several NLP modules, including entity discovery and linking (EDL) and relation detection. Recent success on KB-QA task usually involves complex network structures with sophisticated heuristics. Inspired by a previous work that builds a strong KB-QA baseline, we propose a simple but general neural model composed of fixed-size ordinally forgetting encoding (FOFE) and deep neural networks, called FOFE-net to solve KB-QA problem at different stages. For evaluation, we use two popular KB-QA datasets, SimpleQuestions and WebQSP, and a newly created dataset, FreebaseQA. The experimental results show that FOFE-net performs well on KB-QA subtasks, entity discovery and linking (EDL) and relation detection, and in turn pushing overall KB-QA system to achieve strong results on all datasets. 

\end{abstract}

\begin{figure*}[t]
	\centering
	\includegraphics[width=0.78 \linewidth]{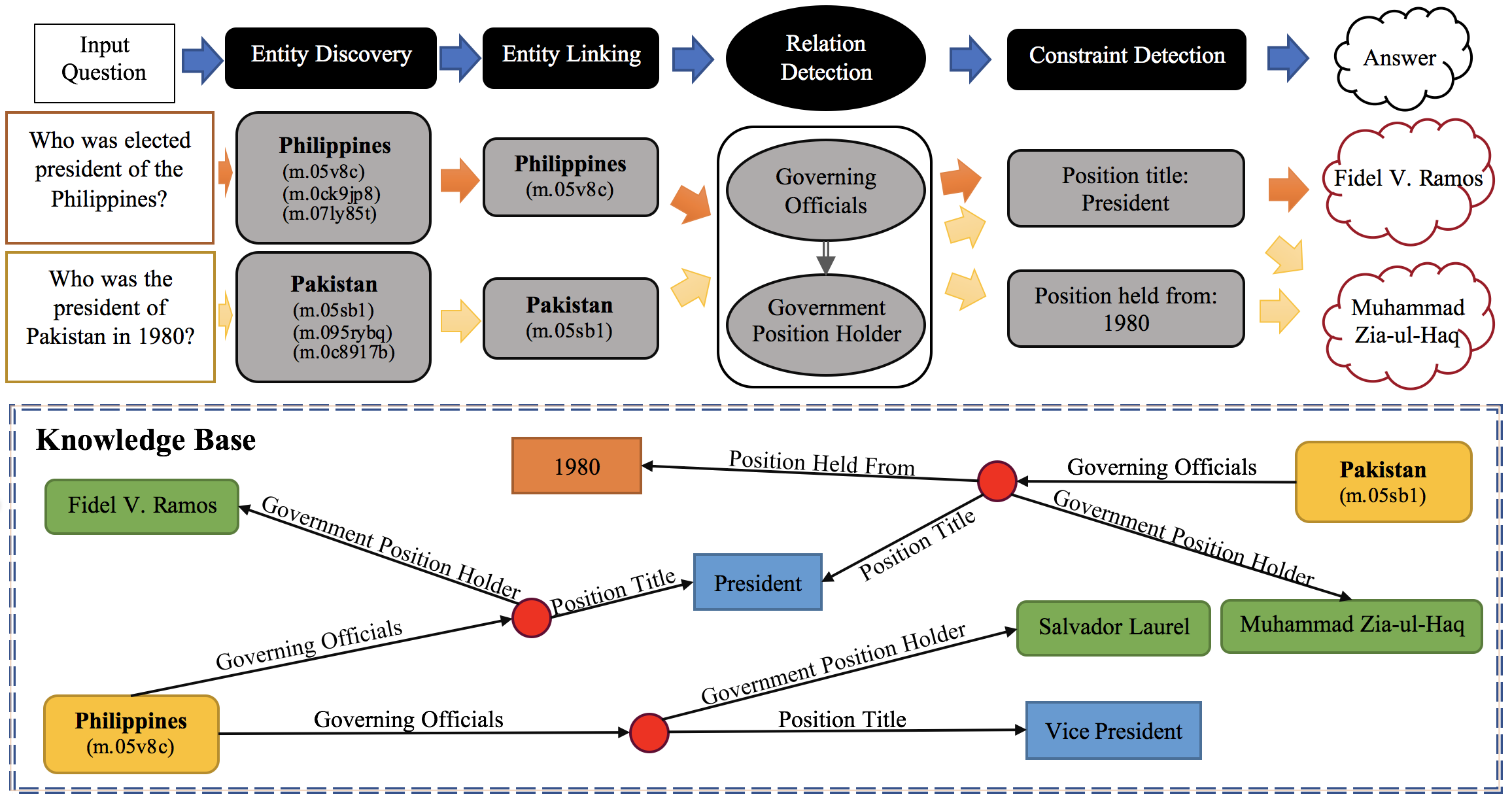}
	\caption{The general pipeline to solve KB-QA problems, adopt in this paper. Given input questions: \textit{Who was elected president of the Philippines?}, the named entity detector should detect \textbf{topic subject mention}: \textit{\{Philippines\}} in question, and entity linker further link this mention to \textbf{subject entity}: \textit{\{m.05v8c\}}. Relation detector detect \textbf{relation / inferential chain} implied in question, which is \textit{\{Governing Officials, Government Position Holder\}} in this case. A set of candidate answers  can be queried by using predicted subject-relation pair. With the \textbf{constraint}: \textit{\{President\}} and \textit{\{Past\}} (not showing in this Figure) get detected, only former presidents of Philippines will be returned as the final answers.}
	\label{KBQA-diagram}
\end{figure*}
\section{Introduction}
\label{sec_1_introduction}
Knowledge bases (KBs) have been intensively studied as a tool to create effective methods to answer factoid questions in natural language processing (NLP). There are several very large-scale knowledge bases \cite{bollacker2008freebase,fabian2007yago,lehmann2015dbpedia} available to solve the KB-QA tasks. In these applications, based on each given question, some models are constructed to retrieve the corresponding supporting facts from the knowledge base to answer the question. 
In many cases, some people have tried to operate the knowledge bases in the similar way as they operate relational databases so that they may come up with a semantic query language, e.g. SPARQL \cite{prud2008sparql}, to enable users to access the correct resources in the knowledge bases. Due to its high demand on user's familiarity with its grammar as well as the structure and vocabulary of knowledge bases, this query language is not very user-friendly. Therefore, much previous work has been done with an attempt to search the resources over knowledge bases via natural languages. For example, 
\citeauthor{D13-1160}~\shortcite{D13-1160} and \citeauthor{semantic-parsing-2015}~\shortcite{semantic-parsing-2015} adopt semantic parsing methods to map each natural-language question to a standard logical form (e.g. lambda calculus) and further translate it to a database query. Some deep learning methods~\cite{D14-1067,10.1007/978-3-662-44848-9_11,DBLP:journals/corr/BordesUCW15,P17-1021} have been adopted to represent questions and facts stored in knowledge bases as low dimensional vectors. Thus, the questions can be answered by computing similarity scores between these vectors. Moreover, some previous works divide the KB-QA task into several NLP subtasks. The answer to each question can be retrieved by performing these subtasks sequentially\cite{P16-1076,C16-1164,P16-1220,P17-1053,DBLP:journals/corr/abs-1804-03317,C18-1277}.

Despite the exciting improvement in the QA task we have seen recently, the neural network structures go increasingly complex. \citeauthor{N18-2047}~\shortcite{N18-2047} build a strong baseline for KB-QA task and prove through experiments that best models at most provide modest gains over their baselines and some of those models show unnecessary complexity. Following their findings, we are curious to look for a simple but stronger architecture that is enabled to compete with the state-of-the-art models on KB-QA task. 

Recently, fixed-size ordinally forgetting encoding (FOFE) method \cite{zhang2015} has been proposed to model variable-length sequences into fixed-size representations in a lossless way. This simple ordinally-forgetting mechanism has been applied to some NLP tasks, e.g. language modeling \cite{zhang2015,D18-1502}, word embeddings \cite{D17-1031}, named entity recognition (NER) \cite{P17-1114}, and have achieved very competitive results. Due to its practical simplicity and technical soundness, we believe it is worthwhile to extend this method to KB-QA task.

To compare our work with other previous ones~\cite{P17-1053,D18-1455}, we choose Freebase~\cite{bollacker2008freebase} as the knowledge base and tackle the KB-QA problem with the following four steps: entity discovery, entity linking, relation detection, and constraint detection. Figure \ref{KBQA-diagram} shows the pipeline of such KB-QA systems. Firstly, name entity detector is responsible for locating topic subject mentions in the question. Secondly, an entity linker will link the detected topic subject mention to the entity node in the knowledge base. Next, relation detector is used to choose the right paths from the entity node to search for the answers. At last, the constraint detector imposes the potential constraints based on the question, which will be used to further prune the found paths and produce a set of answers that may best match the question.

In this paper, we propose a general framework, called FOFE-net, which combines the FOFE encoding method (as a lossless encoding) with a standard feed-forward deep neural networks (DNN) (as a universal approximator) to solve several different subtasks in the above KB-QA pipeline, namely topic subject discovery, entity linking and relation detection. These NLP problems appear to be quite different but we will show that they can be nicely solved with the same model structure of FOFE-net. 
To evaluate the effectiveness of each FOFE-net component in the pipeline as well as the overall KB-QA system, we have used two popular KBQA datasets, i.e. SimpleQuestions~\cite{DBLP:journals/corr/BordesUCW15} and WebQSP \cite{YihRichardsonMeekChangSuh:ACL2016:WebQSP}, and a newly created dataset, FreebaseQA~\cite{N19-696}.\footnote{The FreebaseQA dataset can be found in: \url{https://github.com/infinitecold/FreebaseQA}} Our experimental results have shown that the  FOFE-net models yield very strong performance on each examined subtask for all datasets and our KB-QA system also gives very competitive results on the overall KB-QA task. These results clearly prove the effectiveness of the proposed FOFE-net model on various types of NLP problems. 

\section{Related Work}  

Recently, deep learning methods have achieved great success in many NLP fields, demonstrating its strength in tackling complicated natural language problems. \citeauthor{10.1007/978-3-662-44848-9_11}~\shortcite{10.1007/978-3-662-44848-9_11} firstly introduce neural-network based methods for the KB-QA problem, which maps questions and KB triples to low-dimensional space and the representations of questions. The corresponding answers are made to be more similar throughout the learning process. At the test stage, the best candidate answer will be selected by computing a similarity measure between two projected vectors. \citeauthor{D14-1067}~\shortcite{D14-1067} further improve their method by enriching the representation of answer. The new representation of answers involves question-answer path and all entities associated with the answer. 

Similarly, there is an important line of research work in an end-to-end learning for the KB-QA problem. 
\citeauthor{D16-1166}~\shortcite{D16-1166} propose a character-level encoder-decoder framework enhanced by the attention mechanism for single-relation question answering. This framework uses character-level LSTM and character-level CNNs to encode the questions and the pairs of predicates and entities in the KB respectively. The decoder in this framework is an LSTM with attention mechanism used to generate the topic entity and predicate. \citeauthor{Lukovnikov:2017:NNQ:3038912.3052675}~\shortcite{Lukovnikov:2017:NNQ:3038912.3052675} use a GRU-based model to encode questions and entities on both character and word level and to encode the relations on the word level. The answers might be found by computing the cosine similarities between the question and the entity, and between the question and the predicate.

Our work is closely related to another line of research which tackle the KB-QA problem through multiple steps.
\citeauthor{P17-1053}~\shortcite{P17-1053} propose a KB-QA system built on top of existing entity linkers~\cite{C16-1164,P15-1049}, where they use an entity re-ranker to further improve the entity linking result and a hierarchical BiLSTM model for the relation detection. \citeauthor{C18-1277}~\shortcite{C18-1277} use BiLSTM-CRF to detect the entity mention and extract the question pattern. Fact selection was conducted by computing the cosine similarity between LSTM encoding of question pattern and knowledge base facts.

The contribution of this paper is two-fold. First, the stage-based KB-QA system usually combines the various structure of models in different stages. For example, \citeauthor{P17-1053}~\shortcite{P17-1053} combines S-mart\cite{P15-1049} with hierarchical residual BiLSTM relation detector to solve the KB-QA problem. \citeauthor{D18-1455}~\shortcite{D18-1455} use S-mart to produce a fully-connected graph of the question with edges weighted by the similarity of relation vector and LSTM representation of the question. While in this paper, we only use the same model structure, namely FOFE-net, to solve problems at different stages, which makes the whole pipeline more straightforward and easier to implement. Second, we run our KBQA system on a newly created dataset, FreebaseQA, and provide a detailed analysis and empirical results for it, which may be helpful for the community that has interest in this dataset. 



\section{Backgrounds}

\subsection{Fixed-size Ordinally Forgetting Encoding}

In Zhang et al.~\citeyear{zhang2015}, a new method is proposed to encode any sequence of variable length into a fixed-size representation, called the fixed-size ordinally forgetting encoding (FOFE). It has shown that the FOFE encoding is lossless and unique. 

Let $V$ be a vocabulary set, with each word represented as a one-hot vector. FOFE extends bag-of-words (BoW) by incorporating a forgetting factor to capture positional information. Let $S$ = $w_1, w_2, \cdots, w_T$ represent a sequence of $T$ words, and $e_t$ represent the one-hot vector of the $t$-th word in $S$, where $ 1\le t \le T$. Then the FOFE encoding of the partial sequence $w_1w_2\cdots w_t$ is defined recursively as follows:
\begin{equation}
	\bm{z_t}=
	\begin{cases}
		\bm{0}, & \text{if}\ t = 0 \\
		\alpha \cdot \bm{z_{t - 1}} + \bm{e_t}, & \text{otherwise}
	\end{cases}  
\label{eq:fofe-formula}
\end{equation}

where $\alpha$ denotes the forgetting factor picked between 0 and 1. It is evident that the size of $\bm{z_t}$ is fixed at $|V|$. 

For example, assume a vocabulary $V$ containing the letters A, B and C, with corresponding one-hot vectors being $[1, 0, 0]$, $[0, 1, 0]$ and $[0, 0, 1]$ respectively. Then the FOFE encoding of the sequence ``ABC'' is $[{\alpha}^2, {\alpha}, 1]$ and that of ``ABCBC'' is $[{\alpha}^4, {\alpha} + {\alpha}^3, 1 + {\alpha}^2]$.

It is shown that the encoded word sequences can be unequivocally recovered~\cite{zhang2015}. The uniqueness of the FOFE encoding is guaranteed by the following two theorems:

\begin{theorem}
	For $0 < \alpha \leq 0.5$, FOFE is unique for any countable vocabulary $V$ and any finite value $T$.
\end{theorem}

\begin{theorem}
	For $0.5 < \alpha < 1 $, given any finite value $T$ and any countable vocabulary $V$,
	FOFE is unique almost everywhere, except only a finite set of countable choices of $\alpha$.
\end{theorem}

The uniqueness is not guaranteed when $\alpha$ is chosen to be between 0.5 and 1, the chance of coming across such scenarios is very slim due to quantization errors in the system. Therefore, we can confidently assume that FOFE can uniquely encode any sequence of words of arbitrary length, yielding fixed-size and lossless representations.

\subsection{FOFE-net: combining FOFE with DNNs}

Feedforward deep neural networks (DNNs) use rather simple structure consisting of several fully-connected layers.
These neural networks are known to be powerful as universal approximators \cite{hornik1991}, and they are simpler and faster to train than the more recent variants such as RNNs or CNNs.
Despite the advantages, DNNs require fixed-size inputs, which seems to make it incompetent to deal with complex data types such as sequential data in NLP. 

However, the story may be different if we can combine DNNs with the above FOFE method, where FOFE is used as a frontend encoding method to convert any sequence of words in an NLP task into a fixed-size representation without losing any information and the simple DNNs are used as universal function approximators to map these fixed-size representations into any desirable NLP target labels. This framework, called FOFE-net, may be appealing to many NLP tasks since FOFE is simple and fast and requires neither learning nor feature engineering (except a single hyperparameter), and DNNs are also much easier and faster to manipulate than LSTMs and CNNs. This is particularly important on many NLP tasks when a large corpus is involved in training.
 

\section{FOFE-net for KB-QA}
\label{sec4_FOFE-net_for_KB-QA}
We tackle the KB-QA task with the following four steps: topic subject discovery, entity linking, relation detection, and constraint detection. In this work, we propose to apply the above FOFE-net model to solve each of the sub-tasks in the pipeline. 
Given a question, a FOFE based local detector will examine all word segments in a sentence and generates all candidates along with a probability of being a topic subject mention for the question. After pruning, topic subject mention candidates will be used as keywords to look up the Freebase to generate some candidates of entity nodes in Freebase. Each entity node candidate will be scored by an entity linker, which is a FOFE-net model taking each Freebase node and the corresponding subject mention contexts in the question as input to compute the similarity between them. The results from entity linker are fed to the relation detector, which is implemented as another FOFE-net model to score the similarity between the question and the predicates \footnote{In Freebase, relations between nodes are labeled with certain predicates.} (and paths) from each target Freebase node. Different from the previous works that combine various models in different stages of KB-QA task, our system is mainly built by using the FOFE-net framework. 
The details for each component in our KB-QA system are described in the following subsections.

\subsection{Topic Subject Mention Discovery}
\label{sec_Topic Subject Mention Discovery}
We follow the idea of named entity recognition (NER) in \cite{P17-1114} to discover the potential topic subject mentions from each question in the KB-QA task. Different with regular named entity recognition tasks, detected candidates of topic subject mention has to match any names, alias, or Wikipedia keys in freebase. Those failed to match will be discarded. We then rank the remaining candidates based on their probabilities of being a topic subject mention, as computed by the FOFE-net model. Among all candidates, we only keep the ones whose probability is larger than a threshold  $\theta \; (0 \leq \theta \leq 1)$.


\subsection{Entity Linking} 
\label{Entity Linking}

Once a topic subject mention is detected in a question, it will be used as a keyword to look up entity nodes in Freebase. Usually, each detected subject mention will have multiple matched entity nodes in Freebase (See Figure~\ref{KBQA-diagram}, subject mention: {\textit{Philippines}} has three matched entities: \{\textit{m.05v8c, m.0ck9jp8, m.07ly87t}\}). Thus, an entity linker plays an important role in the process of choosing the correct entity that is relevant to the underlying question. 
Here, we have built the FOFE-net model for entity linking to determine which entity in Freebase should be selected for each detected topic subject mention. To model the relevance between a detected subject mention and each Freebase entity, the 
FOFE-net model takes both Freebase node and the context of detected subject mention in the question as input. In the following, we describe the features used to represent each input. 

\textbf{Features for each subject in the question:} 
Without losing information, each detected subject mention must be represented by its contexts (both left and right) in the given question. Since its left and right contexts may be viewed as a sequence of words, they may be easily encoded as fixed-size FOFE codes.

\begin{figure}[t]
	\centering
	\includegraphics[width=1.0\linewidth]{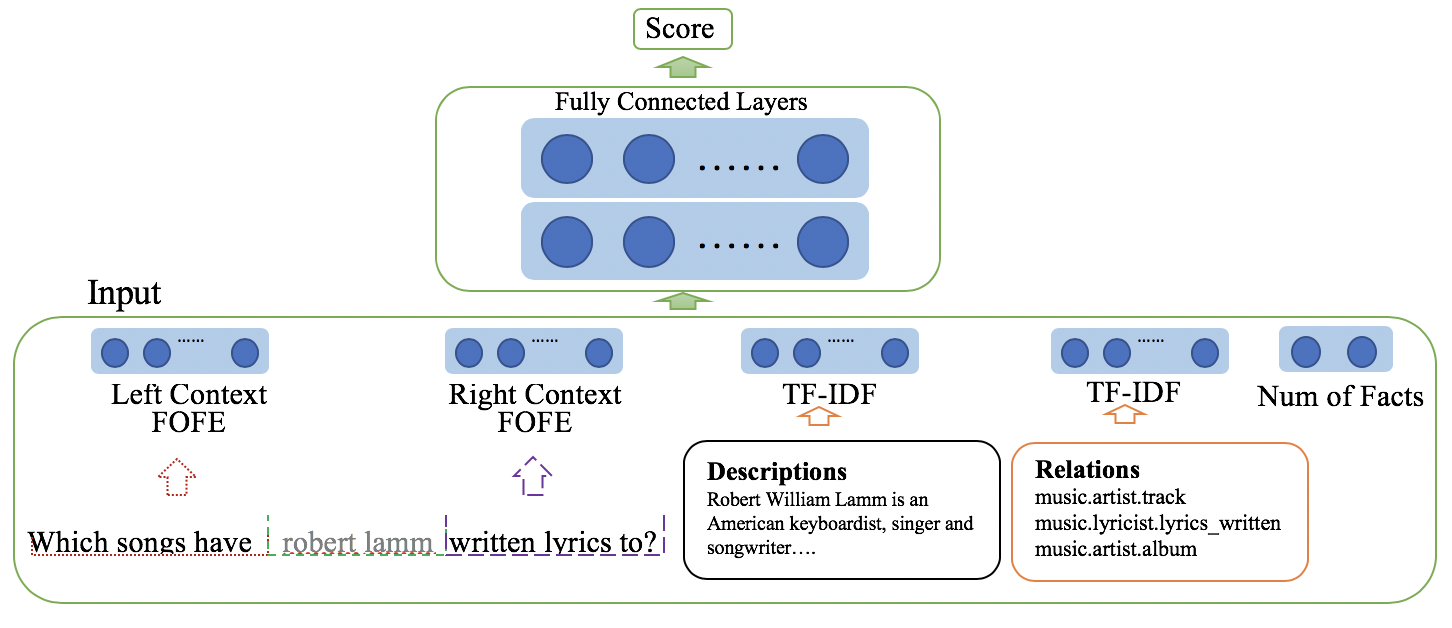}
	\caption{The FOFE-net model for entity linking in the KB-QA pipeline. The detected subject mention in this case is ``\textit{robert lamm}".}
	\label{FOFE-EL-diagram}
\end{figure}
  
As shown by the example in Figure \ref{FOFE-EL-diagram}, the FOFE-net model for entity linking takes the following features for each subject mention in the question:

\begin{itemize}
\item Word-level FOFE codes for left context, e.g. one FOFE code of left context including the detected subject mention:``{\it which songs have robert lamm}", and another FOFE code of left context excluding the subject mention: ``{\it which songs have}".
\item Word-level FOFE codes for right context, e.g. one FOFE code of right context (including the subject mention) as a backward word sequence of ``{\it ? to lyrics written lamm robert}", and another FOFE code for right context excluding the subject mention: ``{\it ? to lyrics written}".
\end{itemize}

\textbf{Features for each Freebase entity node:} 
The FOFE-net also needs to use features to fully depict the information of entity node in Freebase. In this work, As shown in Figure \ref{FOFE-EL-diagram}, we use the following feature for each Freebase node:


\begin{itemize}
	\setlength\itemsep{0.5em}
	\item Number of facts: the total number of associated facts is computed for each node and A hot value is quantized into 10 discrete values and represented as a 10-D one-hot vector \cite{DBLP:journals/corr/LiuLZWJ16}. 
	\item TF-IDF\footnote{The TF-IDF code mentioned in this paper is computed by using TfidfTransformer provided by sklearn. \url{https://scikit-learn.org/stable/modules/generated/sklearn.feature_extraction.text.TfidfTransformer.html}} code for the description of entity node: many Freebase nodes contain a short description of the corresponding entity and a bag of TF-IDF code is computed for the short description. 
	\item TF-IDF code for the relations associated with the entity node: the predicates of all associated relations from the Freebase nodes are all viewed as word sequences and another bag of TF-IDF code is computed for them. 
\end{itemize}

\textbf{Loss Function:} We use a hinge loss function to train our FOFE-net model for entity linking: 
\begin{equation}
L_{linker}=\max \left(0,\gamma+s_{l}\left(q_{c},e^{-}\right)-s_{l}\left(q_{c},e^{+}\right)\right)
\label{eq_el_loss_function_formula}
\end{equation}
where $q_{c}$ denotes question contexts for detected subject mention, {$e^{+}$} denotes the gold subject entity (positive samples), and {$e^{-}$} for all other candidate subject  entities (negative samples), {$\gamma$} is a margin constant.

\textbf{Entity Re-Ranking:} 
To further improve the entity linking, we do the entity re-ranking by combining the linking score with NER score and relation detection score:
\begin{equation}
\begin{aligned}
{\bf s}\left(q, e \right)&= s\left(q, m \right) +s_{l}\left(q_{c}, e\right)
                          &+ \max_{r\in R_{e}} s_{r}\left(q,r\right)  
\label{eq_reranking}
\end{aligned}
\end{equation}
where {$s\left(q, m \right)$} is a similarity score computed by the named entity detector between given question and a topic entity mention.  {$s_{l}\left(q_{c},e\right)$} is a similarity score computed by the entity linker between given question contexts and an entity. {$s_{r}\left(q,r \right)$} is another score computed by the relation detector between a raw question and a relation. {$R_{e}$} denotes the set of all relations associated with entity {$e$}. 
The number of associated facts of an entity is used to break the tie when more than one entity has the same score.


%
%

\subsection{Relation Detection}
\label{sec_Relation Detection}
\begin{figure}[t]
	\centering
	\includegraphics[width=0.95\linewidth]{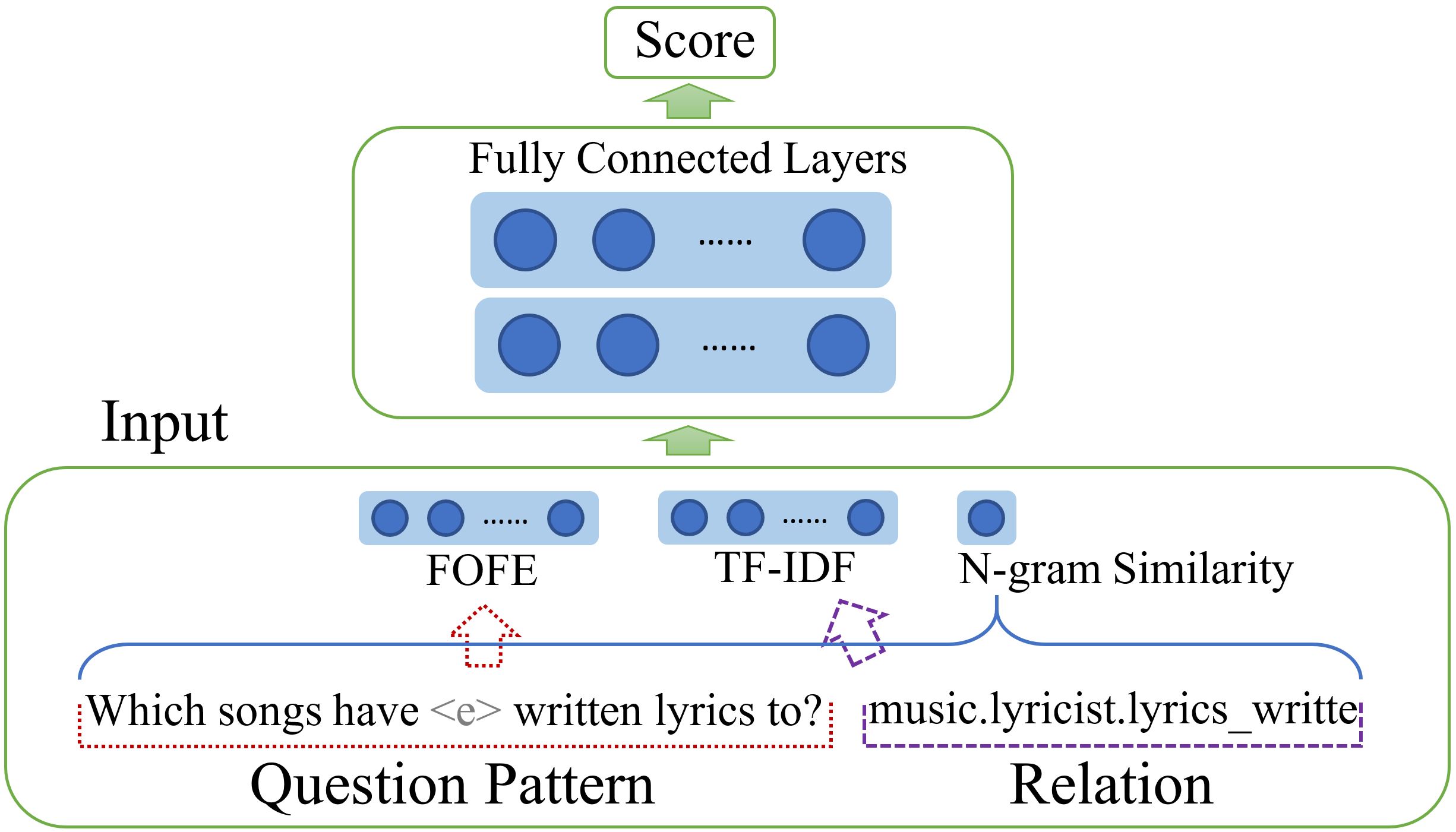}
	\caption{The FOFE model for Relation Detection}
	\label{FOFE-RD-diagram}
\end{figure}

The relation detection module attempts to detect the correct relation that a question refers to. Here we also use the FOFE-net framework to build our relation detector, which takes a question pattern\footnote{Question pattern is formatted by replacing the topic subject mention of the question with a special symbol {$<$}e{$>$}.} and a relation as input to compute a similarity score between them. As shown in Figure \ref{FOFE-RD-diagram}, the following features are fed to the FOFE-net model for relation detection:
%
\begin{description}
    \item[\textbf{Features for Question:}]\
    \begin{itemize}
    	\item Word-level FOFE code for the underlying question pattern viewed as a left-to-right word sequence, e.g. FOFE code of the sequence ``{\it which songs have $<$}e{$>$ written lyrics to ?}" in Figure \ref{FOFE-RD-diagram}. 
    	\item Word-level FOFE code for the question pattern viewed as a right-to-left word sequence, e.g. FOFE code of the sequence ``{\it ? to lyrics written $<$}e{$>$ have songs which}" in Figure \ref{FOFE-RD-diagram}.
    \end{itemize}
    
    \item[\textbf{Features for Relation:}]\
    \begin{itemize}
    	\item A bag of TF-IDF code for relation when the relation predicates are simply viewed as a bag of words, e.g.  the relation of ``{\it music.lyricist.lyrics\_written}" is viewed as a bag of words of ``{\it `music', `lyricist', `lyrics', `written'}".
    	\item Character n-gram feature: the overlap of character n-grams between the question and the relation predicate. 
    \end{itemize} 
\end{description}

\textbf{Loss Function:} 
Similarly, we also use a hinge loss function to train our FOFE-net model for relation detection.
\begin{equation}
L_{rel} =\max \left(0,\gamma+s_{r}\left(q_{p},r^{-}\right)-s_{r}\left(q_{p},r^{+}\right)\right)
\label{eq_rd_loss_function_formula}
\end{equation}
where $q_{p}$ denotes the underlying question pattern, {$r^{+}$} for its gold relation (as positive example), and {$r^{-}$} for other candidate relations associated with the node (as negative samples), and  {$\gamma$} denotes a constant margin parameter.

\subsection{Answer Selection} 
We score each candidate pair of subject entity and relation based on the following formula:
\begin{equation}
s\left(e,r;q\right) = {\bf s}\left(q, e \right) + s_{r}\left(q_{p},r\right)
\label{eq_rel_det}
\end{equation}
where {$s_{r}\left(q_{p},r\right)$} is the similarity score computed by the relation detector between a question pattern and a relation. 

The candidate subject-relation pairs with the top-N highest scores are selected to generate the candidate answers to the question. The score of each candidate answer is the summation of the score of its connected candidate subject-relation pairs. We keep the candidate answers with the highest scores as the final candidate answer set.


\textbf{Constraint Detection} In WebQSP dataset, many of candidate answers of the question can be queried by the same subject-relation pair (which means they may have the same score), while only a few of them are correct because of various kind of constraints implied by the question. Similar to ~\cite{semantic-parsing-2015,YihRichardsonMeekChangSuh:ACL2016:WebQSP,P17-1053}, we also use few pre-defined keywords to detect temporal constraints and ordinal constraints\footnote{For example, \{was, were, did\} and \{first, last\} is used to detect past constraint and ordinal constraint in the question respectively.}. The type entity can be suggested by the predicted inferential chain and detected through text matching.\footnote{We have collected all inferential chains and their associated constraints from the training set.} For example, inferential chain:\{\textit{Governing Official, Government Position Holder}\} suggests several possible constraints: \textit{President},\textit{ Vice President} and \textit{Prime Minister}, etc. Through text matching we can see \textit{President} is the best match constraint to the question: \textit{``who was elected president of the Philippines"}. All detected constraints will be used to prune the answer candidates and produce the final answers to the question.

\section{Experiments}

To evaluate the effectiveness of our proposed FOFE-net models, we have used two popular KB-QA dataset, SimpleQuestions, WebQSP and our newly created dataset, FreebaseQA, in our experiments. We compare our overall QA performance with other systems reported in the published literature \cite{semantic-parsing-2015,YihRichardsonMeekChangSuh:ACL2016:WebQSP,C16-1164,P17-1053,N18-2047,C18-1277,D18-1455}.. Meanwhile, we also investigate the performance of each FOFE-based module in the KB-QA pipeline, i.e. entity linking and relation detection, and compare with other results reported under the same experimental settings \cite{C16-1164,P17-1053,Qiu:2018:HTC:3184558.3186916}.

\subsection{Datasets}

\textbf{SimpleQuestions:} The SimpleQuestions dataset \cite{DBLP:journals/corr/BordesUCW15} is a single-relation KB-QA dataset provided by Facebook along with two extracted subsets of Freebase, i.e. FB2M, and FB5M. Following previous works\cite{DBLP:journals/corr/BordesUCW15,C16-1164,P17-1053,N18-2047,C18-1277}, we use FB2M as the knowledge base and evaluate our system in terms of subject-relation pair accuracy on this benchmark. 

\textbf{WebQSP:} WebQSP dataset \cite{YihRichardsonMeekChangSuh:ACL2016:WebQSP} is a small-scale KB-QA dataset that contains single-relaiton and multi-relation questions. Following previous works~\cite{YihRichardsonMeekChangSuh:ACL2016:WebQSP,P17-1053}, we use entire Freebase for this dataset and choose true accuracy, i.e., a question is considered answered correctly only if the predicted answer set is exactly same as one of the answer sets, and F1 score as the evaluation metric. We use the official evaluation script to evaluate our system on this benchmark.

\textbf{FreebaseQA:} FreebaseQA dataset \cite{N19-696} is a large-scale dataset with 28K unique open-domain factoid questions which collected from triviaQA dataset \cite{P17-1147} and other trivia websites.\footnote{
\begin{tabular}{l}
KnowQuiz (http://www.knowquiz.com)\\[0.5ex]
QuizBalls (http://www.quizballs.com)\\[0.5ex]
QuizZone (https://www.quiz-zone.co.uk)
\end{tabular}
}
All of questions in this dataset have been matched to the Freebase for generating the subject-predicate-object triples and verified by human annotators. We generate a new extract of Freebase (including 182M triples, 16M unique entities) for this dataset and choose true accuracy as the evaluation metric.\footnote{The Freebase extract can be found in: 
\url{https://www.dropbox.com/sh/dkg02j3uwehkt1j/AADqofTAPRA7QpKkhPSW-CJva?dl=0}
}


\begin{table*}[h!]
\begin{center}
\begin{tabular}{l|l|l|c|c|c }
\bf Models&\bf EDL& \bf RD& \bf SQ &\bf WQ  & \bf FQ \\ \hline
\cite{C16-1164}&BiLSTM-CRF&AMPCNN  & 76.4 &-& - \\
\cite{YihRichardsonMeekChangSuh:ACL2016:WebQSP}&S-MART&CNN & - &63.9 & -  \\
\cite{P17-1053}&S-MART&HR-BiLSTM  &77.0&63.0& -  \\ 
\cite{N18-2047}&BiLSTM&BiGRU &75.1&-&-\\
\cite{C18-1277}&BiLSTM-CRF&LSTM&\textbf{80.2}&-&-\\
\cite{D18-1455}&S-MART&LSTM& - &66.7& - \\
\hline
\bf Our work& FOFE-net& FOFE-net& 77.3&\textbf{67.6} & \textbf{37.0}\\
\end{tabular}
\end{center}
\caption{\label{kbqa-end-task} True accuracy (in \%) on the test (eval) sets of SimpleQuestions (SQ), WebQSP (WQ) and FreebaseQA (FQ).}
\end{table*}

\subsection{Experimental Configurations}


We use the same settings as that of \cite{P17-1114} for the topic subject detection model. For the entity linking and relation detecting models, our settings are described as below. 

\textbf{Network structure:} For the feedforward neural networks for entity linking, we choose a structure of 4 fully-connected layers of 1024 nodes with ReLu activation functions on all the datasets. For the network structure for the relation detection model, we use a network of 4 fully-connected layers with ReLu activation function on all the datasets. The number of node in each layers for SimpleQuestions, WebQSP, FreebaseQA is 735, 256, 600 respectively.

\textbf{Embedding Matrices:}  For the entity linker, we use case-insensitive pre-trained word embeddings of 128 dimensions on all the datasets. For our relation detector, we choose case-insensitive pre-trained word embeddings with 128 dimensions on WebQSP and FreebaseQA, and case-insensitive pre-trained word embeddings with 256 dimensions on SimpleQuestions.  Our vocabulary is constituted of the top 100k frequent words from English gigaword~\cite{parker2011english}. 

\textbf{Dropout Rate:} For relation detection models, we have set the dropout rate to 0.24, 0.05, 0.15 on SimpleQuestions, WebQSP and FreebaseQA datasets respectively. For entity linking, we set the dropout rate to 0.15 for all the datasets.

\textbf{Optimizer and Learning rate:} To minimize the loss function of our models, we use the standard SGD (stochastic gradient descent) optimizer with a slow schedule to decay the learning rate. The learning rate is initialized to be 0.01.

\textbf{Forgetting Factor:} we have used {$\alpha = 0.8$} for relation detection on all datasets, and {$\alpha = 0.95$}, {$\alpha = 0.95$} and {$\alpha = 0.8$} for entity linking on SimpleQuestions, WebQSP and FreebaseQA datasets respectively.

\textbf{Training Data Preparation:} To train our FOFE-net models for topic subject discovery, entity linking and relation detection on each specific QA dataset, we only use entire Freebase or its extract to generate negative samples based on this specific QA dataset for model training.  

\subsection{Overall KB-QA Performance}

In Table \ref{kbqa-end-task}, we have listed the overall KB-QA performance of our models on two popular datasets (SimpleQuestions and WebQSP) and compared with other stage-based models reported in the published literature. The column 2 and column 3 list the models that they used for entity discovery and linking~(EDL) and relation detection (RD). We can see the previous works usually combine various models together for the KB-QA task, while our work only use FOFE-net throughout all the stages, which makes the pipeline easy to implement. Our FOFE-net also achieves a very strong performance on the three examined datasets. As shown, our FOFE-net based model is competitive with other CNN, LSTM, regression tree based state-of-the-art model combinations on SimpleQuestions benchmark, and outperform GRAFT-Net~\cite{D18-1455} by 0.9\% on WebQSP in terms of true accuracy. These results prove the effectiveness of our proposed model and suggest that the FOFE-net can be universally applied to other NLP tasks, such as entity discovery and linking (EDL) and relation detection. 

Hao et al.~\cite{C18-1277} report a state-of-the-art result on SimpleQuestions dataset that outperforms other previous works as well as ours by quite large margin (around 3\%). Their model is enhanced by a quite sophisticated rule-based pattern revising method that push their performance from 77.8\% to 80.2\%. Compared to their model, we only use few heuristics and simple functions in combining models at different stages. This simplicity may suffer from some disadvantages on performance while we believe it can make the pipeline more straightforward and easier to adapt to other KB-QA tasks such as WebQSP and FreebaseQA. 

Table \ref{kbqa-end-task} also shows that our FOFE-net model achieves 37.0\% in terms of accuracy on FreebaseQA dataset. Albeit the reasonable performance, it is still worse than that on other datasets by a large margin (30-40 \%), which demonstrates that the FreebaseQA dataset is more challenging than SimpleQuestions and WebQSP.   

\begin{table*}[h!]
\centering
 \begin{tabular}{l | c | c| c | c} 
 \hhline{=====}
  \multicolumn{5}{c}{ \bf SimpleQuestions}\\ \cline{0-4}
  \hhline{=====}
   &\multicolumn{4}{c}{\bf Top-K}\\ \cline{2-5}
 \textbf{ Models} & \textbf{1} & \textbf{10} &\textbf{20}& \textbf{50} \\ [0.5ex] 
 \hline
AMPCNN \cite{C16-1164}&72.7&86.9&88.4&90.2\\
HR-BiLSTM \cite{P17-1053}&79.0&89.5&90.9&92.5\\

HTTED \cite{Qiu:2018:HTC:3184558.3186916} &81.1&91.7&93.4&\textbf{95.1}\\
 {\bf Our FOFE-net} &\textbf{82.2}&\textbf{92.5}&\textbf{93.6}&94.7\\
  \hhline{=====}
 \multicolumn{5}{c}{ \bf WebQSP}\\ \cline{0-4}
 \hhline{=====}
    &\multicolumn{4}{c}{\bf Top-K}\\ \cline{2-5}
\textbf{ Models} & \textbf{1} & \textbf{2} &\textbf{3}& \textbf{5} \\ [0.5ex] 
 \hline
S-Mart \cite{P15-1049}&88.0&\textbf{93.3}&\textbf{94.7}&\textbf{96.0}\\
 {\bf Our FOFE-net} &\textbf{89.1}&93.2&93.7&94.1\\
 \hhline{=====}
  \multicolumn{5}{c}{ \bf FreebaseQA}\\ \cline{0-4}
  \hhline{=====}
     &\multicolumn{4}{c}{\bf Top-K}\\ \cline{2-5}
 \textbf{ Models} & \textbf{1} & \textbf{2} &\textbf{3}& \textbf{5}\\ [0.5ex] 
 \hline
 {\bf Our FOFE-net} &52.4&70.7&79.6&85.7\\
 \hline
 \end{tabular}
 \caption{\label{el-b} The accuracy (in \%) of various entity linking models on the test (eval) set of SimpleQuestions, WebQSP and FreebaseQA in the number of candidates kept in the list.}
\end{table*}


\begin{table*}[h!]
\begin{center}
\begin{tabular}{l|c|c|c}
\hline 
&\multicolumn{3}{c}{\bf Accuracy}\\ \cline{2-4}
\bf Models & \bf SQ  & \bf WQ &\bf FQ\\ \hline
AMPCNN \cite{C16-1164} & 91.3 & -&- \\
HR-BiLSTM \cite{P17-1053} & 93.3 & 82.53&- \\\hline
{\bf Our FOFE-net} & \textbf{93.3} & \textbf{83.26}&76.3\\
\hline
\end{tabular}
\end{center}
\caption{\label{rel-det} Comparison of relation detection accuracy (in \%) on the test (eval) sets of SimpleQuestions (SQ), WebQSP (WQ) and FreebaseQA (FQ).}
\end{table*}

\subsection{Entity Linking Results}

To investigate the performance of FOFE-net in entity linking task, we have compared it with three state-of-the-art entity linkers \cite{C16-1164,P17-1053,Qiu:2018:HTC:3184558.3186916} on SimpleQuestions and S-Mart~\cite{P15-1049} entity linker on WebQSP benchmark. From the results in Table \ref{el-b}, we can see that our entity linker achieves a very competitive result in entity linking subtask and outperforms the state-of-the-art models in terms of top 1 accuracy.\footnote{In sec.~\ref{sec_Topic Subject Mention Discovery}, we choose a large {$\theta$} to prevent entity linker from being overwhelmed by a large number of candidate entities, which affects the entity linking result when K goes large.}

According to the results in Table \ref{el-b}, we can also see that entity linking on FreebaseQA is more difficult than it is on WebQSP and SimpleQuestions. To analyze the reason, we calculate the average ratio of number of topic entity mention candidates to number of gold topic entity mentions for questions on test (eval) set of three datasets. The results show that that ratio on SimpleQuestions, WebQSP and FreebaseQA is 8.0, 4.4, 11.6 respectively, which have linear positive correlation with the average length of questions in these datasets. Intuitively, the more redundant entity mention candidates each question has, the more difficult for the entity linker to make the right prediction. This may explain why our entity linker performs worse on Freebase in comparison with other datasets.



\subsection{Relation Detection Results}

To investigate the FOFE-net performance on relation detection task, we have compared it with the best relation detector reported in the published literature \cite{C16-1164,P17-1053} on SimpleQuestions and WebQSP benchmark. As shown in Table \ref{rel-det}, our FOFE-net based relation detector yields very competitive results on both dataset, which ties HR-BiLSTM on SimpleQuestions and slightly outperform it on WebQSP by 0.73\%.

From the Table \ref{rel-det}, we can also see that the result our FOFE-net model achieves on FreebaseQA is quite reasonable, even the number is still worse than it is on either WebQSP or SimpleQuestions dataset.

\section{Conclusion}

FOFE, a simple fixed-size encoding method, has been successfully applied to many NLP tasks. In this paper, we have further proposed to combine this method with simple feedforward neural networks to solve several sub-tasks in the KB-QA pipeline, including topic subject discovery, entity linking and relation detection. Our experimental results have shown that this simple framework, called the FOFE-net, works well in all examined sub-tasks, and in turn pushing our overall KB-QA system to achieve competitive results on two popular KB-QA datasets and a newly created dataset.

\bibliography{naaclhlt2019}
\bibliographystyle{acl_natbib}

\end{document}